\documentclass[letterpaper, 10 pt, conference]{ieeeconf}
\IEEEoverridecommandlockouts                              
\usepackage{times}
\usepackage{epsfig}
\usepackage{graphicx}
\usepackage{amsmath}
\usepackage{amssymb}

\usepackage{graphicx}
\usepackage{epsfig}
\usepackage{epic,eepic}
\usepackage{times}
\usepackage{mathptmx}
\usepackage{amsfonts}
\usepackage{amsmath}
\usepackage{cite}
\usepackage{color}

\usepackage{times}

\definecolor{red}{rgb}{1,0,0}
\definecolor{green}{rgb}{0,1,0}
\definecolor{blue}{rgb}{0,0,1}
\definecolor{violet}{rgb}{1,0,1}
\definecolor{cyan}{cmyk}{1,0,0,0}
\definecolor{magenta}{cmyk}{0,1,0,0}
\definecolor{yellow}{cmyk}{0,0,1,0}

\definecolor{white}{rgb}{1,1,1}

\usepackage{jumoline}

\newcommand{\CO}[1]{}

\newcommand{\CommentOut}[1]{}

 \newcommand{\editage}[1]{}

\begin{document}

\newcommand{\FIG}[3]{
\begin{minipage}[b]{#1cm}
\begin{center}
\includegraphics[width=#1cm]{#2}\\
{\scriptsize #3}
\end{center}
\end{minipage}
}

\newcommand{\FIGU}[3]{
\begin{minipage}[b]{#1cm}
\begin{center}
\includegraphics[width=#1cm,angle=180]{#2}\\
{\scriptsize #3}
\end{center}
\end{minipage}
}

\newcommand{\FIGm}[3]{
\begin{minipage}[b]{#1cm}
\begin{center}
\includegraphics[width=#1cm]{#2}\\
{\scriptsize #3}
\end{center}
\end{minipage}
}

\newcommand{\FIGR}[3]{
\begin{minipage}[b]{#1cm}
\begin{center}
\includegraphics[angle=-90,width=#1cm]{#2}
\\
{\scriptsize #3}
\vspace*{1mm}
\end{center}
\end{minipage}
}

\newcommand{\FIGRpng}[5]{
\begin{minipage}[b]{#1cm}
\begin{center}
\includegraphics[bb=0 0 #4 #5, angle=-90,clip,width=#1cm]{#2}\vspace*{1mm}
\\
{\scriptsize #3}
\vspace*{1mm}
\end{center}
\end{minipage}
}

\newcommand{\FIGpng}[5]{
\begin{minipage}[b]{#1cm}
\begin{center}
\includegraphics[bb=0 0 #4 #5, clip, width=#1cm]{#2}\vspace*{-1mm}\\
{\scriptsize #3}
\vspace*{1mm}
\end{center}
\end{minipage}
}

\newcommand{\FIGtpng}[5]{
\begin{minipage}[t]{#1cm}
\begin{center}
\includegraphics[bb=0 0 #4 #5, clip,width=#1cm]{#2}\vspace*{1mm}
\\
{\scriptsize #3}
\vspace*{1mm}
\end{center}
\end{minipage}
}

\newcommand{\FIGRt}[3]{
\begin{minipage}[t]{#1cm}
\begin{center}
\includegraphics[angle=-90,clip,width=#1cm]{#2}\vspace*{1mm}
\\
{\scriptsize #3}
\vspace*{1mm}
\end{center}
\end{minipage}
}

\newcommand{\FIGRm}[3]{
\begin{minipage}[b]{#1cm}
\begin{center}
\includegraphics[angle=-90,clip,width=#1cm]{#2}\vspace*{0mm}
\\
{\scriptsize #3}
\vspace*{1mm}
\end{center}
\end{minipage}
}

\newcommand{\FIGC}[5]{
\begin{minipage}[b]{#1cm}
\begin{center}
\includegraphics[width=#2cm,height=#3cm]{#4}~$\Longrightarrow$\vspace*{0mm}
\\
{\scriptsize #5}
\vspace*{8mm}
\end{center}
\end{minipage}
}

\newcommand{\FIGf}[3]{
\begin{minipage}[b]{#1cm}
\begin{center}
\fbox{\includegraphics[width=#1cm]{#2}}\vspace*{0.5mm}\\
{\scriptsize #3}
\end{center}
\end{minipage}
}



\newcommand{\acprPaperID}{25}





\onecolumn

\title{\LARGE \bf
Mining Minimal Map-Segments for Visual Place Classifiers
}
\author{%
Tanaka Kanji 
\thanks{Our work has been supported in part by 
JSPS KAKENHI 
Grant-in-Aid 
for Scientific Research (C) 26330297, and (C) 17K00361.}
\thanks{K. Tanaka is with Faculty of Engineering, University of Fukui, Japan. 
{\tt\small tnkknj@u-fukui.ac.jp}}
\thanks{We would like to express our sincere gratitude to Kousuke Yamaguchi, Koji Takeda, and Ryohei Yamamoto for development of deep learning architectures, and initial investigation on VPR tasks on the dataset, which helped us to focus on our map segmentation project.}
}
\maketitle

\begin{abstract}
In visual place recognition (VPR),
map segmentation (MS)
is a preprocessing technique used to partition
a given view-sequence map
into place classes
(i.e., map segments)
so that
each class has
good
place-specific training images
for a visual place classifier (VPC).
Existing approaches to
MS
implicitly/explicitly
suppose
that map segments have a certain size,
or individual map segments are balanced in size.
However,
recent VPR systems
showed that
very small important map segments
(minimal map segments)
often suffice for VPC,
and the remaining large unimportant portion
of the map 
should be discarded
to minimize map maintenance cost.
Here,
a new MS algorithm
that
can mine minimal map segments
from 
a large view-sequence map
is presented.
To solve the inherently NP hard problem,
MS
is formulated
as a video-segmentation problem and
the efficient point-trajectory based paradigm of video segmentation is used.
The proposed map representation
was implemented 
with three types of VPC:
deep convolutional neural network,
bag-of-words,
and object class detector,
and
each was integrated 
into
a Monte Carlo localization algorithm (MCL)
within
a topometric VPR framework.
Experiments
using the publicly available NCLT dataset
thoroughly
investigate the efficacy of
MS
in terms of
VPR performance.
\end{abstract}

\section{Introduction}

\newcommand{\figA}{
\begin{figure}[t]
\begin{center}
\FIG{15}{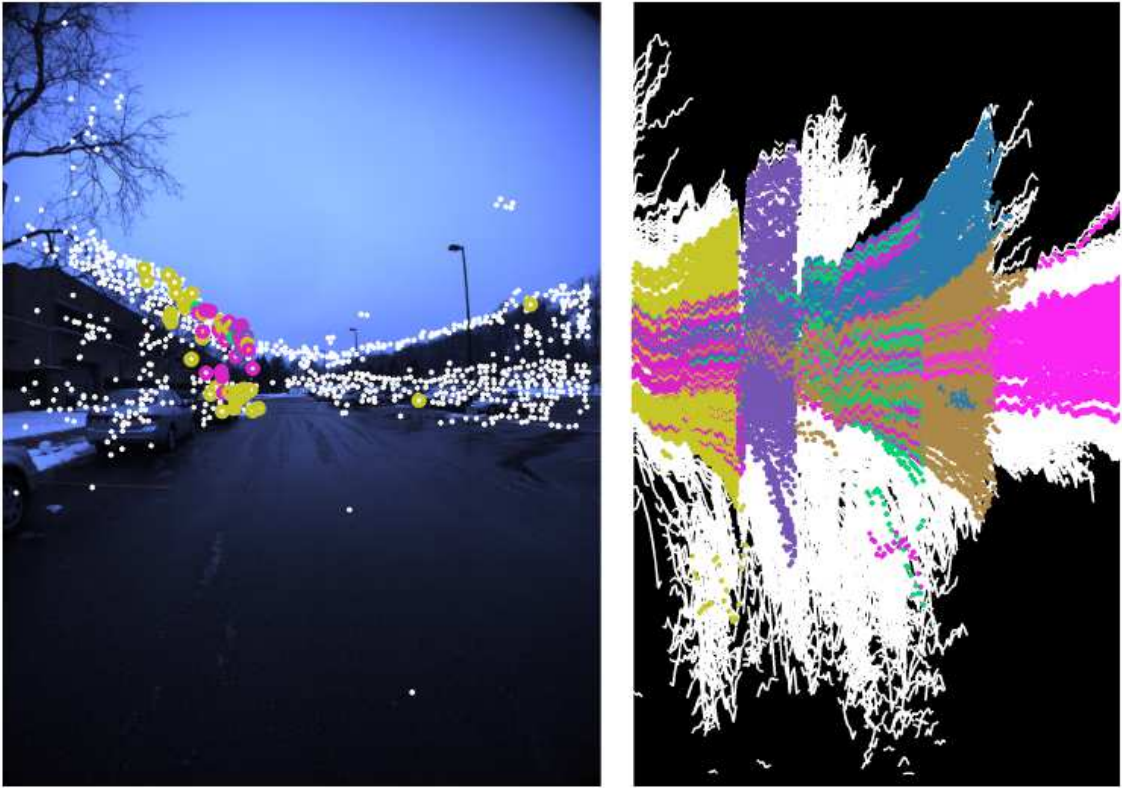}{}
\caption{%
View-image sequence map
segmented
into
important minimal map segments (colored points/curves)
and
the remaining large unimportant portion (white points/curves),
which are very unbalanced in size:
a segmentation result
of
the map (i.e., KLT point trajectories) in $x$-$y$-$t$ spatio-temporal coordinate
is visualized by 
projecting
it onto the $x$-$y$ image plane (left panel),
and
onto the $y$-$t$ plane (right panel)
---
different colors
represent
different map segments.
  }\label{fig:A}
\end{center}
\end{figure}
}

\newcommand{\tabA}{
\begin{table}
\begin{center}
\caption{VPR Performance in Top-$X$ accuracy [\%].}\label{tab:A}
\begin{tabular}{r|rrrrr|}
$X$ & 10  & 20  & 50  & 100  & 200 \\ \hline 
BOW  & 66.5 & 74.2 & 83.4 & 89.8 & {\bf 95.7} \\
MMM-BOW  & 54.2 & 64.6 & 73.5 & 80.6 & {\bf 91.7} \\ 
CNN  & 74.8 & 79.4 & 88.0 & 93.8 & {\bf 96.6} \\
MMM-CNN  & 44.6 & 54.5 & 72.9 & 86.2 & {\bf 95.1} \\ 
MMM-OCD  & 8.9 & 16.6 & 35.7 & 57.5 & 83.4 \\ \hline 
\end{tabular}
\end{center}
\end{table}
}

\newcommand{\figD}{
\begin{figure}[t]
\begin{center}
\scriptsize
\FIG{15}{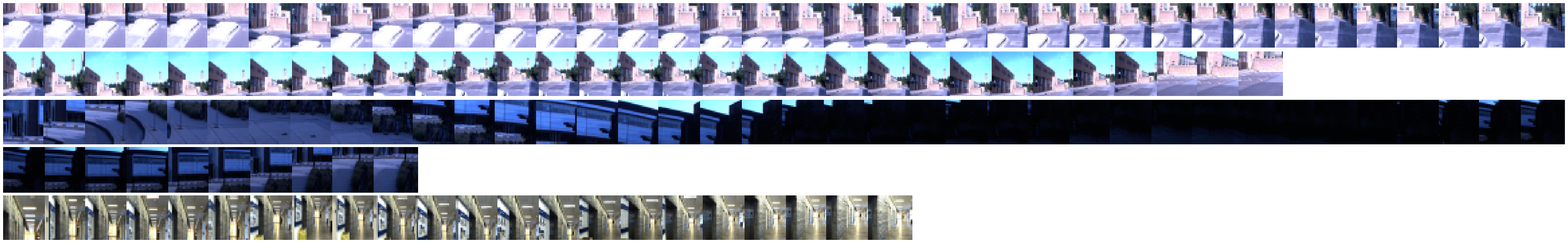}{}\\
a\\
\FIG{15}{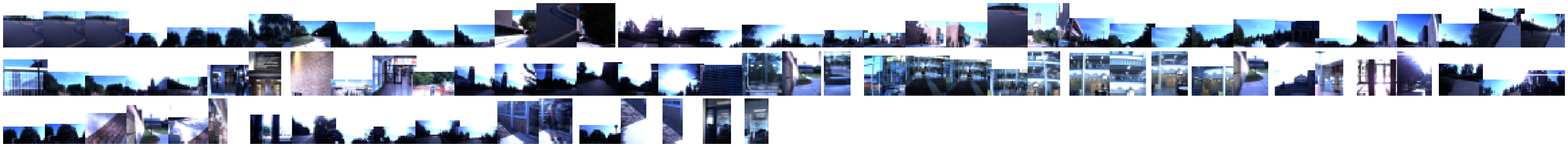}{}\\
b
\caption{%
Examples of subimages cropped by MS
for the season ``2012/8/4":
(a) inner-class variation and
(b) interclass variation (random representative images from the 94 classes).
  }\label{fig:D}
\end{center}
\end{figure}
}

\newcommand{\figE}{
\begin{figure}[t]
\begin{center}
\FIG{15}{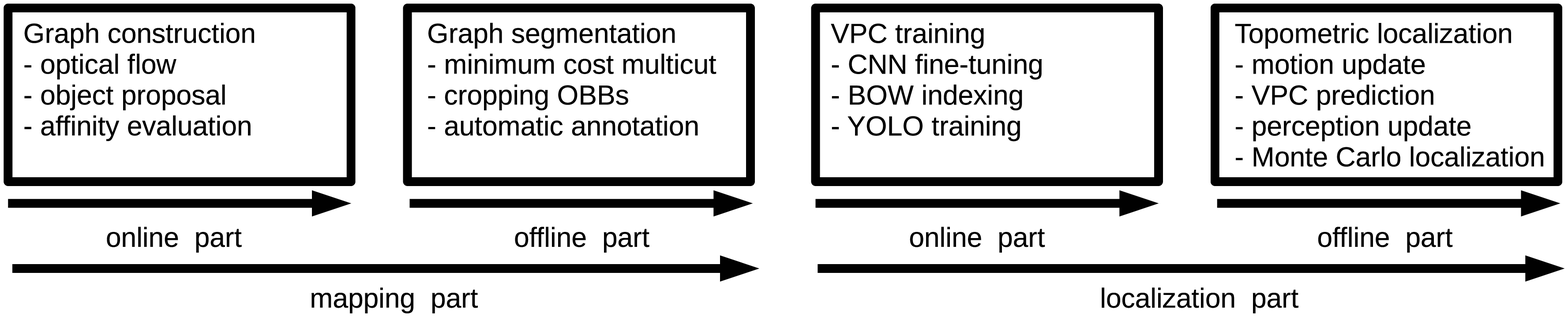}{}
\caption{System architecture.}\label{fig:E}
\end{center}
\end{figure}
}

\newcommand{\figG}{
\begin{figure}[t]
\begin{center}
\FIG{15}{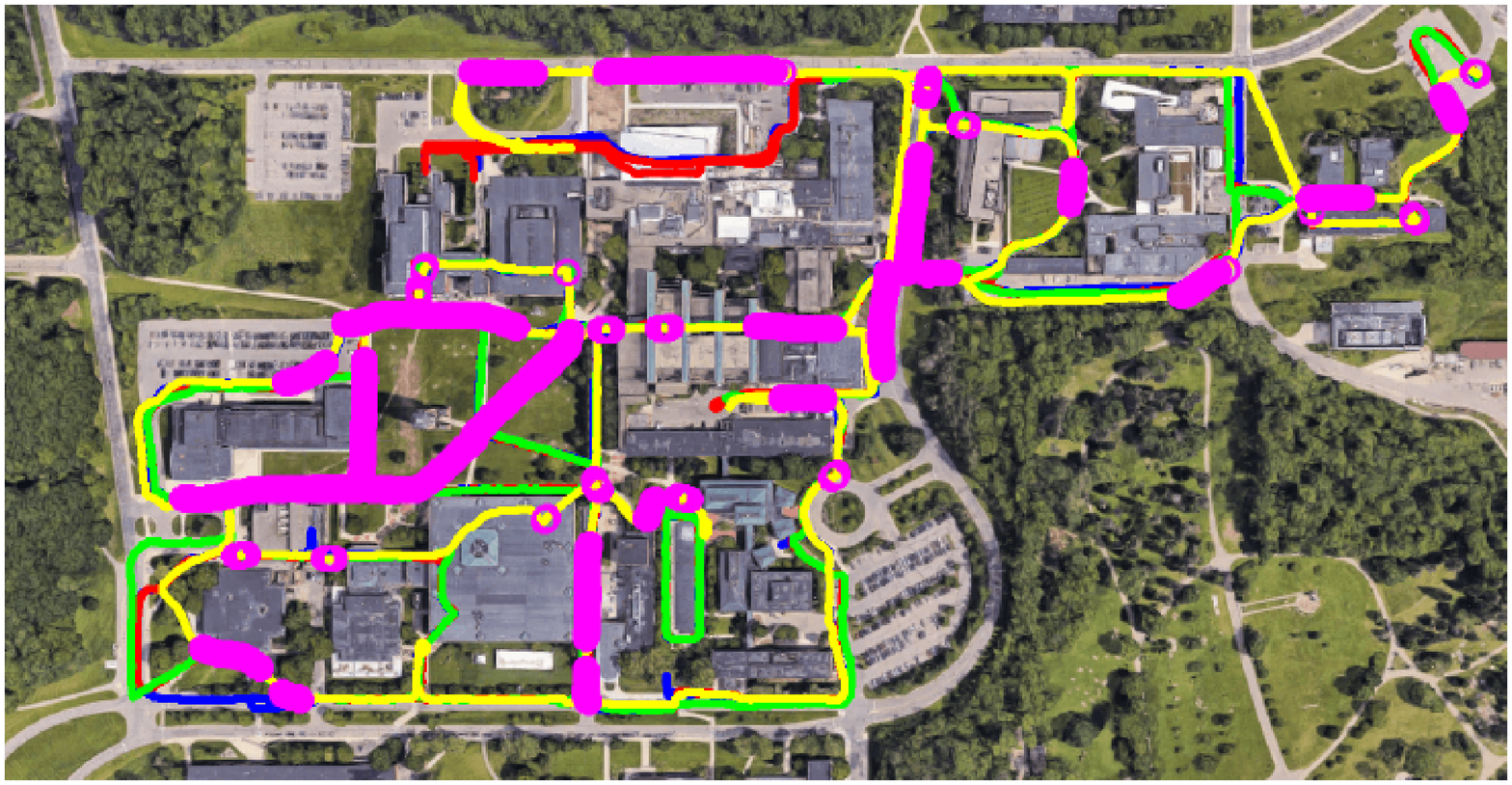}{}
\caption{%
Experimental environments, robot trajectories, and map segments:
curves with blue, red, green, and yellow represent trajectories
for seasons ``2012/1/22", ``2012/3/31", ``2012/8/4", and ``2012/11/17"
---
purple big circles represent viewpoints that belong to map segments from season ``2012/8/4".
}\label{fig:G}
\end{center}
\end{figure}
}

In visual place recognition (VPR),
map segmentation (MS)
is a preprocessing method used to partition
a given view-sequence map
into place classes
(i.e., map segments),
so that
each class has
good
place-specific training images
for a visual place classifier (VPC).
This MS problem has become an important research topic in
the community of robotic mapping and localization
\cite{ref3, ref4, ref5},
because of the growing interest
in VPCs
(e.g., deep VPC \cite{survey16vpr}). 
Current approaches to
MS
implicitly/explicitly
suppose
that map segments have a certain size,
or individual map segments are balanced in size \cite{ref15}.
However,
recent VPR systems \cite{ref2} 
have shown that
very small important map segments
(minimal map segments)
often suffice for VPC,
and the remaining large, unimportant portion
of the map 
should be discarded
to minimize map maintenance cost.
Such
important and unimportant
map segments are very unbalanced in size,
which
makes it more difficult
to apply the existing MS approaches.

In this work,
a new MS algorithm
is presented
that
can mine minimal map segments
from 
a large view-sequence map (Fig. \ref{fig:A}).
To solve the inherently NP hard problem,
MS
was formulated
as a video-segmentation problem,
and the efficient point-trajectory based paradigm of video segmentation
was utilized.
The MS task consists of online and offline sub-tasks.
Online,
a mapper robot
navigates the target environment
while
incrementally constructing
a point trajectory graph
in real-time
by integrating
per-frame
optical flows \cite{KLT} 
and
object proposals \cite{YOLO}. 
Offline,
it aims to partition the graph into
important minimal map segments
and the remaining,
large unimportant portion.
Thus,
the resulting
segments should be very unbalanced in size.
The predominant
approach,
spectral clustering-based point trajectory segmentation \cite{K15},
relies on the assumption of balanced segment sizes,
and thus not suited for such unbalanced size settings.
To address this issue,
we adopt
the minimum cost multicut algorithm
derived from the field of
image/video segmentation \cite{K7, K25, K}. 
This made it possible
to 
specify (not only positive, but also) negative affinities between
important and unimportant segments
and
to avoid
joining a small segment into a
large neighboring segment.

\figA

Furthermore,
the mined map segments
(i.e., place-classes)
with VPCs
were implemented,
and the VPR performance
was investigated thoroughly.
Specifically,
three case studies
were conducted
on
three different VPCs,
by plugging each
into
a Monte Carlo localization algorithm \cite{MCL} 
within 
a topometric localization framework \cite{TML}. 
In the first case,
a deep convolutional neural network (CNN) \cite{VGG} 
was introduced as a VPC
by treating each map segment as
(a set of) place-specific training images.
In the second case,
state-of-the-art
bag-of-words (BOW) -based loop closure detection \cite{IBOWLCD} 
was
introduced as an appearance-based VPC
by treating each map segment as place-specific visual words.
In the third case,
an object class detector (OCD) technique \cite{YOLO} 
was introduced as a segment class detector
by treating each map segment
as
class-specific training images.
For training,
a method
is proposed
for self-supervised learning,
by which
the object bounding boxes (OBBs)
of the training images
can be
automatically annotated.
The developed VPC systems were evaluated
in challenging cross-season VPR scenarios \cite{cs1}, 
using the publicly available NCLT dataset \cite{NCLT}. 
The experimental results showed
that
the proposed approach
frequently achieved
comparable VPR performance 
to the state-of-the-art approaches
even though
it only used minimal map segments.

\section{Approach}\label{sec:approach}

Fig. \ref{fig:E}
illustrates the overview of our VPR system.
As shown,
the MS system
consists of
online and offline sub-tasks.
In the online sub-task,
the mapper robot
incrementally constructs
a point trajectory graph (Section \ref{sec:define}),
while it navigates the target environment
by integrating
per-frame
optical flows
and
object proposals (Section \ref{sec:online}).
In the latter sub-task,
it partitions the constructed graph
into small important segments (i.e., place classes)
and unimportant large segments to discard
(Section \ref{sec:segment}).
Because the proposed map representation
maintains topological information
only for such a small portion of the map,
the topometric localization framework \cite{TML} 
was adopted
rather than metric 
or topological localization (Section \ref{sec:vpr}).

\subsection{Graph-Based Map Segmentation}\label{sec:define}

In our graph-based MS framework,
a given view-sequence map
is 
interpreted to a point trajectory graph
$G=(V,E)$.
$V$
is a set of vertices,
each of which
represents
a point trajectory
or
an object proposal.
$E$
is a set of 
weighted edges,
each of which 
represents
affinity between the vertex pair.
Then,
the MS problem is formulated as
partitioning the graph
into an optimal number of segments
by minimizing 
overall cost
in terms of edge weight $c_e$:
\begin{equation}
y^* = \arg \min_{y\in Y} \sum_{e \in E} c_e y_e. \label{eqn:objective}
\end{equation}
$Y$
is a set of all possible multicuts:
$\{0,1\}^E$.
Note that 
the number of map segments
$C$
is obtained as
the number of connected components in the resulting multicut $y^*$.

Vertices of
$V$
are based on 
spatial-temporal curves
called point trajectories.
This design choice
is motivated by the fact
that
such 
point trajectories
are
reliably
estimated by
optical flow techniques (Section \ref{sec:online}).
Moreover,
recent research on VPR \cite{ref2}, 
motion segmentation \cite{ref31},
and novelty detection \cite{ref39} 
showed
that such a point trajectory
is often
a stable part of the environment (e.g., landmarks). 

Weights of edges
$E$
can be either 
positive or negative.
Positive edges
are
those edges
that should be joined
in graph partitioning.
Negative edges
are those
that should be cut.
As shown in video segmentation literature,
positive edges are useful for 
segmenting out important foreground objects (e.g., landmarks) from the background (e.g., Fig.\ref{fig:D}a top).
However,
recent works on VPR like \cite{ref2} 
showed that
not all parts of a foreground object are equally salient (e.g., Fig.\ref{fig:D}a middle)
and
many scenes have no foreground object (e.g., Fig.\ref{fig:D}a bottom).
In such general cases,
negative edges
are useful for
dividing
foreground/background regions
into small salient subregions (i.e., landmarks)
and the remaining large non-salient subregions.
Consequently,
positive and negative edges
are useful
and maintained in our approach (Section \ref{sec:online}).

\subsection{Incremental Graph Construction}\label{sec:online}

\figE

The mapper robot
incrementally builds
the point trajectory graph
by incorporating real-time image measurements 
during the navigation.
This is
realized
by two sub-tasks.
One is 
incremental estimation of graph vertices.
The other is 
incremental evaluation
of the affinity between each vertex pair.

For trajectory estimation,
the KLT tracker \cite{KLT}
was adopted. 
KLT
is one of most widely used
optical flow estimation techniques
in robotics \cite{ref32} 
and autonomous driving \cite{ref33}. 
Formally,
$N=1,500$
features
at the initial frame
are created
and
tracked over successive frames.
When
a few
$N' (\ll N)$
features
are lost
(because of occlusions, limited field-of-view, or background clutters),
new 
$N'$
features
are initialized,
and
the lost $N'$ features
are replaced by the new ones.

For affinity evaluation,
a semantic cue
from
the object class detector
is used.
In preliminary studies \cite{kanji2015unsupervised, 
ref36, 
ref37}, 
the effectiveness of
the other possible cues:
color, spatial, and motion cues,
was also evaluated
in the affinity evaluation.
Color is an effective cue for foreground object detection \cite{kanji2015unsupervised}, 
but 
it is often
not invariant
and
not consistent
under varying outdoor illumination conditions.
Spatial cues
or distance between object locations
is useful for image segmentation \cite{ref36},  
but
determining an appropriate threshold
according to
individual object sizes
is inherently a difficult problem,
which
makes the graph segmentation unstable.
Motion
is
an effective cue
for
motion segmentation \cite{ref37}, 
but
it is difficult
to segment
the
relative motions of static objects
(i.e., map segments)
in our application domain of MS.
On the other hand,
object bounding boxes (OBBs)
from general purpose object class detectors \cite{BING,YOLO,EDGEBOX} 
provide
stable
semantic cues
to join or separate
segments.
Moreover,
one can expect
OBBs
to provide
an additional spatial cue
---
that is,
two point trajectories
belonging to the same OBB
can be considered to be spatially close to each other.
Such an OBB based semantic cue
was recently used 
to enhance point feature matching
in a different context of 3D reconstruction \cite{icra19keypoint}.

Based on the above consideration,
semantic OBB measurements
from
the object class detector
are used
for affinity evaluation.
More formally,
an edge
between
a newly arrived OBB (node)
and
a point trajectory (node)
that belongs to the OBB is inserted.
Whether
a trajectory being tracked
belongs to a newly arrived OBB,
can be easily checked
using a few simple arithmetic and logical operations.
For simplicity,
the affinity value
of every edge is set to 1.
Although
typical object detectors
provide not only OBBs
but also prediction of their object classes,
it was decided not to use this additional semantic cue,
because
even state-of-the-art object detectors
frequently
fail to predict correct class labels
especially for nearly-unseen objects.
After the decision,
a tiny YOLO detector \cite{TINY}
was chosen, 
because it
provides
rapid class-agnostic OBB detection.

\subsection{Graph Partitioning}\label{sec:segment}

The remaining problem is how to
solve
the optimization problem
in Eq.\ref{eqn:objective}
to partition 
the graph
into
the optimal number of segments.
A natural optimization approach 
would be
applying spectral clustering \cite{K9,K13,K7,K30,K26} 
or its recent variants such as 
multi-label graph-cut \cite{K32} 
or
unbalanced energy \cite{K19}. 
Although 
these methods
can easily
specify which trajectories should belong to the same segment,
they do not
specify
which should be separated.
Therefore,
they are not suited for
unbalanced size sub-graphs.
However,
recently developed
minimum-cost multicut approaches
(e.g.,
image segmentation \cite{K1,K2,K20}, 
pixel graphs \cite{K22}, 
motion trajectory segmentation \cite{K}, 
and pedestrian tracking \cite{K37}) 
can explicitly
represent
not only positive but
also negative affinities between edges,
which act as a repulsive force between segments \cite{K15}. 
More formally,
we adopt
the heuristics
in \cite{K4} 
that
partitions the graph
with complexity $O(n^2 \log n)$.
Importantly,
the sizes of segments
can be controlled by
subtracting a bias $c_o$
from the edge weights
(i.e., $c_e\leftarrow$$c_e-c_o$)
prior to MS.
In this study,
this bias $c_o$
was set to
the 20\% highest weight over all the edges in the graph,
which yields approximately 1/5 smaller map segments than the original map,
as demonstrated in experiments in Section \ref{sec:exp}.

\subsection{VPR with Mined Minimal Map Segments (MMMs)}\label{sec:vpr}

VPCs were implemented using the mined map segments (i.e., place classes)
and each was integrated into the MCL \cite{MCL}
within the topometric localization framework \cite{TML}.
For simplicity,
the drift-free motion model \cite{ref34}, 
was assumed
and the number of particles $N=D/D_o$
was set
according to the map size in terms of travel distance $D=100$ m
normalized by a constant $D_o=1.0$.
At the initialization step $t=0$,
the $N$ particles $\{p_i\}_{i=1}^N$ are uniformly distributed over the entire map trajectory with travel distance $D$,
as in previous research \cite{metric2}.
At each time step,
the MCL processings of 
motion update and perception update
are performed.
The confidence score
$\Delta L(p)$ for
ego-location hypothesis $p$
output by the VPC
is normalized
to ensure
$\sum_p \Delta L(p) = 1$,
and used to update the likelihood 
$L(p)$
in the form:
$L(p)\leftarrow L(p)+\Delta L(p)$.

VPR performance is evaluated by 
a ranked list of the ego-location hypotheses
at the goal location
with
respect to
the ground truth (i.e., GPS).
In the spirit of Monte Carlo simulation,
the robot navigation with MCL
is iterated for 
$N'=D/100$
different start locations
separated by 100m,
and
the resulting $N'$ ranked lists
at the goal locations
are summarized 
into
the Top-$X$ accuracy performance index ($X=$10, 20, 50, 100, and 200),
using
nonmaximum suppression (NMS) \cite{ref38} 
to 
obtain a less redundant hypothesis set.
That is,
outputting an ego-location hypothesis $p$ is suppressed
if
a higher-ranked hypothesis $p'$
already occupies
the location: $|p-p'|<10$ [m].

For
VPC,
three different methods
were implemented:
deep CNN,
BOW,
and OCD.

The
CNN
method
formulates the VPC as
a classification task,
and employs
a deep VPC with
the Vgg16 CNN architecture \cite{VGG}. 
In this case,
each map segment is treated as
place-specific train images.
For learning and prediction,
map images are 
resized to $256$$\times$$256$
before
being input to the CNN.
In addition,
a different setting
is also considered
where
the above training images are cropped
by the bounding boxes of the class-specific point trajectories
before being resized.
This variant is termed
``PartCNN"
and
was also tested,
as described in Section \ref{sec:exp}.
To avoid instability in training,
an image from a map segment
is not considered as the member of the training set
if 
its
bounding box (before being resized)
is smaller than 100 pixels in width or in height,
for both the CNN and PartCNN.

The
BOW
method
formulates
the VPC
as a BOW image retrieval task,
and it employs
the state-of-the-art BOW
loop closure detection framework from previous work \cite{IBOWLCD}.
In this case,
each map segment
is treated as
place-specific visual words.
This BOW framework
is based on ORB features \cite{ORB}, 
the TF-IDF scoring scheme \cite{sivic2003video},
and the ratio-test \cite{lowe2004distinctive}
with novel incremental vocabulary \cite{IBOW},
and the
retrieval outputs are further refined by
the island-based place clustering \cite{galvez2012bags}
and
RANSAC-based geometric verification.
In previous work,
we studied this BOW framework
in a different context
of simultaneous mapping and localization \cite{ref36}. 
In the current study,
it was necessary to modify slightly the framework
and
implement
mapping (i.e., learning) and
localization (i.e., prediction)
as two separate processes.
As in the PartCNN method,
a variant, PartBOW,
was also considered and tested,
where
cropped train subimages are used in place of
non-cropped original images.

The OCD method
formulates the VPC
as an alternative image retrieval task
using bag of segment classes (in place of BOW) as the cue,
and it employs
the state-of-the-art
object class detector from a prior study \cite{YOLO}.
In this case,
each map segment is treated as 
class-specific training images.
Unlike the pre-trained generic object detector (in Section \ref{sec:online}),
a new detector is fine-tuned on the training images
to
predict place class directly.
The
fine-tuning
task
requires
annotations 
in the form of OBBs.
In the present approach,
such an OBB
can be approximated by the bounding box
of the class-specific point trajectories projected onto the image plane.
These OBBs
can be automatically computed as the byproduct of our graph-based MS.
As in the CNN and PartCNN methods,
training images with very small bounding boxes
are discarded.
Once the detector is trained,
the bag of segment classes
is predicted
and then
used to index/search an inverted file.
The original annotated class labels
could be used to index,
but
it was found that
the predicted class labels
work better
in practice.
Such a
segment-class-based indexing
is an extremely compact
$\log C$ bit
($\in [6, 8]$ in the experiments)
representation
for a subimage.

\section{Experiments}\label{sec:exp}

The MS approach was demonstrated
using the publicly available NCLT dataset \cite{NCLT}.
The main goal of
the experiments was to evaluate
the MS algorithm
in terms of
the performance
of VPR using MMMs.

\subsection{Dataset and Performance Index}

The NCLT dataset is a large-scale,
long-term autonomy dataset for robotics research collected at the University of Michigan's North Campus by a Segway vehicle robotic platform.
Recently,
this dataset has been widely used in robotics communities as an experimental benchmark for various tasks,
such as map-merging \cite{ref40}. 
The data used in the current study
include view image sequences along a vehicle's trajectories acquired by the front-facing
camera of the Ladybug3 with GPS.
Specifically,
four datasets
---
``2012/1/22 (WI)", ``2012/3/31 (SP)", ``2012/8/4 (SU)", and ``2012/11/17 (AU)"
---
collected across four different seasons
were used.
The
image size was 1232$\times$1616.
Fig. \ref{fig:G} shows 
the experimental environment
and examples of viewpoint trajectories
in the dataset.

VPR performance was evaluated
by Top-$X$ accuracy [\%]
according to the viewpoint hypotheses at the goal location in MCL.
A correct hypothesis is defined as a viewpoint hypothesis
whose distance to the ground truth GPS viewpoint is nearer than 10m.
For evaluation,
test view-sequences
whose overlap ratios to seen viewpoints were lower than 80\%
or
whose overlap ratios to minimal map segments were lower than 10\%
were discarded
from the test set.

\subsection{Implementation Details}

For KLT, features in previous work \cite{shi1994good} are employed with maxCorners=200, qualityLevel=0.05, minDistance=5.0.
The number of features per frame was set to 1500.
For YOLO \cite{YOLO},
the dimension of the training network and the batch size were modified to 256$\times$256$\times$3, and to 32.
The initial learning rate was 0.001 and reduced on plataue (by the factor 0.1, patience 3).
Early stopping (patience 10) was used.
For Vgg16 \cite{VGG}, the batch size, the number of epochs, and validation samples were set to 32, 10, and 10, respectively.
For MCL \cite{MCL}, the number of hypotheses was proportional to map size, and approximately $10^4$.
For MS \cite{ref15},
the approach of 
equal-length subsequences 
is used as a default baseline method.
The proposed graph-construction algorithm
yields
a bipartile graph rather than a complete graph.
That is,
no edge exist between
point trajectory vertex pairs.
This is an important property
because
dealing with
the huge number (e.g., $(10^6)^2$) of trajectory vertex pairs 
is computationally intractable.
As a result of MS,
the number of classes
was 131, 127, 94, and 119
for the season
``WI", ``SP", ``SU", and ``AU",
respectively,
while
the number of trajectory vertices per class were
288.8$\pm$367.4,
381.4$\pm$291.6,
229.5$\pm$183.0,
and
200.1$\pm$114.29,
respectively.

\figG

\figD

\subsection{MS Results}

Fig. \ref{fig:D}
shows example results
for the proposed MS method
for training set ``SU".
It can be seen
that
coherent and salient
map segments
were successfully obtained.
MS was successful,
even when
there was no foreground object (e.g., Fig.\ref{fig:D}a bottom),
and
even when
not all parts of foreground objects were salient (e.g., Fig.\ref{fig:D}a middle).

The map-maintenance cost
is described
by two quantities 
$R_i$ and $R_p$.
$R_i$
is the number of map images that belong to any mined map segment,
normalized by the number of the entire map images.
$R_p$
is the number of pixels that belong to any mined map segment,
normalized by the number of pixels in the map images.
Whether a pixel on an image belongs to a map segment
was checked by using
the bounding box of the map segment projected onto the image.
The results of evaluation were
$R_i$ 
=
26.9\%, 29.0\%, 21.2\%, and 28.2\%,
and
$R_p$
= 4.93\%, 6.20\%, 3.74\%, and 7.02\%
for seasons ``WI", ``SP", ``SU", and ``AU", respectively.
For those map images that belong to any map segment,
the mean and standard deviation of the number of pixels per image
[\%]
were
18.2$\pm$18.7,
21.4$\pm$22.5,
17.6$\pm$17.6,
and
24.9$\pm$24.7,
for seasons ``WI", ``SP", ``SU", and ``AU", respectively.

Although
the MS was performed independently for different seasons,
the mined map segments (e.g., landmarks)
were expected to be invariant across seasons to some extent.
To investigate
the amount of such invariance,
the
similarity 
of the map segments
between a query and a reference seasons
was
evaluated.
For simplicity,
each map-segment
was represented
as a set of discretized viewpoints 
and
the Jaccard index 
of the set
was evaluated
between different seasons.
For discretization,
a grid of cells with size $10\times 10$ m
was employed
and
a cell ID was used as the discretized viewpoint.
The similarity
between each query segment
and its most similar reference segment
for all 12 possible query-reference-season-pairs
was investigated.
The result of the evaluation
is summarized as follows.
(1) The ratio of query segments with zero similarity values
ranges
from 0.468 to 0.771 for the 12 season pairs.
(2) The maximum, mean and medium similarities
of the other segment pairs with non-zero similarity values ranged
from 0.601 to 1,
0.202 to 0.367,
and
0.149 to 0.333
for the 12 season pairs.

\subsection{MS Performance}

Tab. \ref{tab:A}
shows
results for cross-season VPR
for all the 12 paired
live/map seasons.
In the table,
BOW, CNN, OCD, PartBOW, and PartCNN
are
the VPR systems
using different types of VPC method,
as described
in Section \ref{sec:vpr}.
MMM-$X$ ($X\in$ \{BOW, CNN, OCD\})
are
VPR results
using only MMMs.
As can be seen,
the proposed MMM-BOW and MMM-CNN
had
comparable performance to
those of 
BOW and CNN,
which require the entire original map.
In these experiments,
the proposed framework
successfully captured the
stable part of the map (e.g., Fig. \ref{fig:D}a)
and
these stable map segments
acted as useful landmarks in VPR.
However,
MMM-OCD
could not perform well
in the current experiments.
This was mainly 
the result of the 
high mis-detection rate
of the object class detector.
In addition,
the cross-season scenario
(i.e., trained and tested in different seasons)
was very challenging
even for the state-of-the-art YOLO detector.
Based on the above results,
it is concluded
that the
proposed approach
frequently
yielded
comparable VPR performance
to the state-of-the-art VPR methods
even though
it used only MMMs.

\tabA

\section{Conclusions and Future Works}

A point trajectory based MS framework
ppfor
mining minimal map segments
from
a view-sequence map
was proposed.
A
computationally tractable
minimum-cost multicut-based MS algorithm
was proposed,
that can 
specify (not only positive but also) negative affinities between
important and unimportant segments,
to avoid
joining a small segment into a
large neighboring segment.
It can take
advantage of
optical flows
and
object proposals,
which
increases
computational efficiency
in the online task of
constructing a point trajectory graph.
Our approach has shown to be effective
in providing
good
place-specific train images
for
a VPC.

In future work,
it is planned to expand the range of
MS
and
include different map models
(e.g., bird's eye view map \cite{ref3},
and 3D maps \cite{ref10}),
which
we were not considered here.
Additionally,
the point trajectory model
could represent
a wide range of
segmentation cues,
such as
color, spatial, and motion
cues
as discussed in Section \ref{sec:online}.
It is planned to
explore a more general framework
for
combining different segmentation cues
to improve
robustness against noises
(e.g., occlusions, limited field of view, or background clutter)
and other general-purpose image/video segmentation techniques \cite{seg1}.
The proposed MS framework
automatically finds
a compact set of landmarks (i.e., minimal map segments);
however, the compactness
might be improved
by introducing landmark selection techniques \cite{ref27}.

\bibliographystyle{IEEEtran}
\bibliography{lel19}

\begin{thebibliography}{10}
\providecommand{\url}[1]{#1}
\csname url@rmstyle\endcsname
\providecommand{\newblock}{\relax}
\providecommand{\bibinfo}[2]{#2}
\providecommand\BIBentrySTDinterwordspacing{\spaceskip=0pt\relax}
\providecommand\BIBentryALTinterwordstretchfactor{4}
\providecommand\BIBentryALTinterwordspacing{\spaceskip=\fontdimen2\font plus
\BIBentryALTinterwordstretchfactor\fontdimen3\font minus
  \fontdimen4\font\relax}
\providecommand\BIBforeignlanguage[2]{{%
\expandafter\ifx\csname l@#1\endcsname\relax
\typeout{** WARNING: IEEEtran.bst: No hyphenation pattern has been}%
\typeout{** loaded for the language `#1'. Using the pattern for}%
\typeout{** the default language instead.}%
\else
\language=\csname l@#1\endcsname
\fi
#2}}

\bibitem{ref3}
M.~Mielle, M.~Magnusson, and A.~J. Lilienthal, ``A method to segment maps from
  different modalities using free space layout maoris: map of ripples
  segmentation,'' in \emph{2018 IEEE International Conference on Robotics and
  Automation (ICRA)}.\hskip 1em plus 0.5em minus 0.4em\relax IEEE, 2018, pp.
  4993--4999.

\bibitem{ref4}
M.~Liu, F.~Colas, F.~Pomerleau, and R.~Siegwart, ``A markov semi-supervised
  clustering approach and its application in topological map extraction,'' in
  \emph{2012 IEEE/RSJ International Conference on Intelligent Robots and
  Systems}.\hskip 1em plus 0.5em minus 0.4em\relax IEEE, 2012, pp. 4743--4748.

\bibitem{ref5}
E.~A. Topp and H.~I. Christensen, ``Detecting structural ambiguities and
  transitions during a guided tour,'' in \emph{2008 IEEE International
  Conference on Robotics and Automation}.\hskip 1em plus 0.5em minus
  0.4em\relax IEEE, 2008, pp. 2564--2570.

\bibitem{survey16vpr}
S.~M. Lowry, N.~S{\"{u}}nderhauf, P.~Newman, J.~J. Leonard, D.~D. Cox, P.~I.
  Corke, and M.~J. Milford, ``Visual place recognition: {A} survey,''
  \emph{{IEEE} Trans. Robotics}, vol.~32, no.~1, pp. 1--19, 2016.

\bibitem{ref15}
T.~Hiroki and K.~Tanaka, ``Long-term knowledge distillation of visual place
  classifiers,'' in \emph{2019 22st International Conference on Intelligent
  Transportation Systems (ITSC)}.\hskip 1em plus 0.5em minus 0.4em\relax IEEE,
  2019.

\bibitem{ref2}
T.~Naseer, G.~L. Oliveira, T.~Brox, and W.~Burgard, ``Semantics-aware visual
  localization under challenging perceptual conditions,'' in \emph{2017 IEEE
  International Conference on Robotics and Automation (ICRA)}.\hskip 1em plus
  0.5em minus 0.4em\relax IEEE, 2017, pp. 2614--2620.

\bibitem{KLT}
B.~D. Lucas, T.~Kanade, \emph{et~al.}, ``An iterative image registration
  technique with an application to stereo vision,'' 1981.

\bibitem{YOLO}
J.~Redmon and A.~Farhadi, ``Yolov3: An incremental improvement,'' \emph{arXiv
  preprint arXiv:1804.02767}, 2018.

\bibitem{K15}
K.~Fragkiadaki and J.~Shi, ``Detection free tracking: Exploiting motion and
  topology for segmenting and tracking under entanglement,'' in \emph{CVPR
  2011}.\hskip 1em plus 0.5em minus 0.4em\relax IEEE, 2011, pp. 2073--2080.

\bibitem{K7}
T.~Brox and J.~Malik, ``Object segmentation by long term analysis of point
  trajectories,'' in \emph{European conference on computer vision}.\hskip 1em
  plus 0.5em minus 0.4em\relax Springer, 2010, pp. 282--295.

\bibitem{K25}
J.~Lezama, K.~Alahari, J.~Sivic, and I.~Laptev, ``Track to the future:
  Spatio-temporal video segmentation with long-range motion cues,'' in
  \emph{CVPR 2011}.\hskip 1em plus 0.5em minus 0.4em\relax IEEE, 2011, pp.
  3369--3376.

\bibitem{K}
M.~{Keuper}, B.~{Andres}, and T.~{Brox}, ``Motion trajectory segmentation via
  minimum cost multicuts,'' in \emph{2015 IEEE International Conference on
  Computer Vision (ICCV)}, Dec 2015, pp. 3271--3279.

\bibitem{MCL}
F.~Dellaert, D.~Fox, W.~Burgard, and S.~Thrun, ``Monte carlo localization for
  mobile robots,'' in \emph{ICRA}, vol.~2, 1999, pp. 1322--1328.

\bibitem{TML}
H.~Badino, D.~Huber, and T.~Kanade, ``Real-time topometric localization,'' in
  \emph{2012 IEEE International Conference on Robotics and Automation}.\hskip
  1em plus 0.5em minus 0.4em\relax IEEE, 2012, pp. 1635--1642.

\bibitem{VGG}
K.~Simonyan and A.~Zisserman, ``Very deep convolutional networks for
  large-scale image recognition,'' \emph{CoRR}, vol. abs/1409.1556, 2014.

\bibitem{IBOWLCD}
E.~Garcia-Fidalgo and A.~Ortiz, ``ibow-lcd: An appearance-based loop-closure
  detection approach using incremental bags of binary words,'' \emph{IEEE
  Robotics and Automation Letters}, vol.~3, no.~4, pp. 3051--3057, 2018.

\bibitem{cs1}
M.~J. Milford and G.~F. Wyeth, ``Seqslam: Visual route-based navigation for
  sunny summer days and stormy winter nights,'' in \emph{2012 IEEE
  International Conference on Robotics and Automation}.\hskip 1em plus 0.5em
  minus 0.4em\relax IEEE, 2012, pp. 1643--1649.

\bibitem{NCLT}
N.~Carlevaris-Bianco, A.~K. Ushani, and R.~M. Eustice, ``University of michigan
  north campus long-term vision and lidar dataset,'' \emph{The International
  Journal of Robotics Research}, vol.~35, no.~9, pp. 1023--1035, 2016.

\bibitem{ref31}
A.~O{\v{s}}ep, W.~Mehner, P.~Voigtlaender, and B.~Leibe, ``Track, then decide:
  Category-agnostic vision-based multi-object tracking,'' in \emph{2018 IEEE
  International Conference on Robotics and Automation (ICRA)}.\hskip 1em plus
  0.5em minus 0.4em\relax IEEE, 2018, pp. 1--8.

\bibitem{ref39}
\BIBentryALTinterwordspacing
A.~Osep, P.~Voigtlaender, J.~Luiten, S.~Breuers, and B.~Leibe, ``Large-scale
  object mining for object discovery from unlabeled video,'' in
  \emph{International Conference on Robotics and Automation, {ICRA} 2019,
  Montreal, QC, Canada, May 20-24, 2019}, 2019, pp. 5502--5508. [Online].
  Available: \url{https://doi.org/10.1109/ICRA.2019.8793683}
\BIBentrySTDinterwordspacing

\bibitem{ref32}
H.~Ho, C.~De~Wagter, B.~Remes, and G.~De~Croon, ``Optical-flow based
  self-supervised learning of obstacle appearance applied to mav landing,''
  \emph{Robotics and Autonomous Systems}, vol. 100, pp. 78--94, 2018.

\bibitem{ref33}
M.~Siam, H.~Mahgoub, M.~Zahran, S.~Yogamani, M.~Jagersand, and A.~El-Sallab,
  ``Modnet: Motion and appearance based moving object detection network for
  autonomous driving,'' in \emph{2018 21st International Conference on
  Intelligent Transportation Systems (ITSC)}.\hskip 1em plus 0.5em minus
  0.4em\relax IEEE, 2018, pp. 2859--2864.

\bibitem{kanji2015unsupervised}
K.~Tanaka, ``Unsupervised part-based scene modeling for visual robot
  localization,'' in \emph{Robotics and Automation (ICRA), 2015 IEEE
  International Conference on}.\hskip 1em plus 0.5em minus 0.4em\relax IEEE,
  2015, pp. 6359--6365.

\bibitem{ref36}
R.~Yamamoto, K.~Tanaka, and K.~Takeda, ``Invariant spatial information for
  loop-closure detection,'' in \emph{2019 16th International Conference on
  Machine Vision Applications (MVA)}.\hskip 1em plus 0.5em minus 0.4em\relax
  IEEE, 2019, pp. 1--6.

\bibitem{ref37}
T.~Murase, K.~Tanaka, and A.~Takayama, ``Change detection with global viewpoint
  localization,'' in \emph{2017 4th IAPR Asian Conference on Pattern
  Recognition (ACPR)}.\hskip 1em plus 0.5em minus 0.4em\relax IEEE, 2017, pp.
  31--36.

\bibitem{BING}
M.-M. Cheng, Z.~Zhang, W.-Y. Lin, and P.~Torr, ``Bing: Binarized normed
  gradients for objectness estimation at 300fps,'' in \emph{Proceedings of the
  IEEE conference on computer vision and pattern recognition}, 2014, pp.
  3286--3293.

\bibitem{EDGEBOX}
C.~L. Zitnick and P.~Doll\'ar, ``Edge boxes: Locating object proposals from
  edges,'' in \emph{ECCV}, 2014.

\bibitem{icra19keypoint}
X.~Huang, Z.~Dai, W.~Chen, L.~He, and H.~Zhang, ``Improving keypoint matching
  using a landmark-based image representation,'' in \emph{2019 International
  Conference on Robotics and Automation (ICRA)}.\hskip 1em plus 0.5em minus
  0.4em\relax IEEE, 2019, pp. 1281--1287.

\bibitem{TINY}
J.~Redmon, S.~Divvala, R.~Girshick, and A.~Farhadi, ``You only look once:
  Unified, real-time object detection,'' in \emph{Proceedings of the IEEE
  conference on computer vision and pattern recognition}, 2016, pp. 779--788.

\bibitem{K9}
A.~M. Cheriyadat and R.~J. Radke, ``Non-negative matrix factorization of
  partial track data for motion segmentation,'' in \emph{2009 IEEE 12th
  International Conference on Computer Vision}.\hskip 1em plus 0.5em minus
  0.4em\relax IEEE, 2009, pp. 865--872.

\bibitem{K13}
R.~Dragon, B.~Rosenhahn, and J.~Ostermann, ``Multi-scale clustering of
  frame-to-frame correspondences for motion segmentation,'' in \emph{European
  Conference on Computer Vision}.\hskip 1em plus 0.5em minus 0.4em\relax
  Springer, 2012, pp. 445--458.

\bibitem{K30}
P.~Ochs, J.~Malik, and T.~Brox, ``Segmentation of moving objects by long term
  video analysis,'' \emph{IEEE transactions on pattern analysis and machine
  intelligence}, vol.~36, no.~6, pp. 1187--1200, 2013.

\bibitem{K26}
Z.~Li, J.~Guo, L.-F. Cheong, and S.~Zhiying~Zhou, ``Perspective motion
  segmentation via collaborative clustering,'' in \emph{Proceedings of the IEEE
  International Conference on Computer Vision}, 2013, pp. 1369--1376.

\bibitem{K32}
H.~Rahmati, R.~Dragon, O.~M. Aamo, L.~Van~Gool, and L.~Adde, ``Motion
  segmentation with weak labeling priors,'' in \emph{German Conference on
  Pattern Recognition}.\hskip 1em plus 0.5em minus 0.4em\relax Springer, 2014,
  pp. 159--171.

\bibitem{K19}
P.~Ji, H.~Li, M.~Salzmann, and Y.~Dai, ``Robust motion segmentation with
  unknown correspondences,'' in \emph{European conference on computer
  vision}.\hskip 1em plus 0.5em minus 0.4em\relax Springer, 2014, pp. 204--219.

\bibitem{K1}
B.~Andres, J.~H. Kappes, T.~Beier, U.~K{\"o}the, and F.~A. Hamprecht,
  ``Probabilistic image segmentation with closedness constraints,'' in
  \emph{2011 International Conference on Computer Vision}.\hskip 1em plus 0.5em
  minus 0.4em\relax IEEE, 2011, pp. 2611--2618.

\bibitem{K2}
B.~Andres, T.~Kroeger, K.~L. Briggman, W.~Denk, N.~Korogod, G.~Knott,
  U.~Koethe, and F.~A. Hamprecht, ``Globally optimal closed-surface
  segmentation for connectomics,'' in \emph{European Conference on Computer
  Vision}.\hskip 1em plus 0.5em minus 0.4em\relax Springer, 2012, pp. 778--791.

\bibitem{K20}
J.~H. Kappes, M.~Speth, B.~Andres, G.~Reinelt, and C.~Schn, ``Globally optimal
  image partitioning by multicuts,'' in \emph{International Workshop on Energy
  Minimization Methods in Computer Vision and Pattern Recognition}.\hskip 1em
  plus 0.5em minus 0.4em\relax Springer, 2011, pp. 31--44.

\bibitem{K22}
M.~Keuper, E.~Levinkov, N.~Bonneel, G.~Lavou{\'e}, T.~Brox, and B.~Andres,
  ``Efficient decomposition of image and mesh graphs by lifted multicuts,'' in
  \emph{Proceedings of the IEEE International Conference on Computer Vision},
  2015, pp. 1751--1759.

\bibitem{K37}
S.~{Tang}, B.~{Andres}, M.~{Andriluka}, and B.~{Schiele}, ``Subgraph
  decomposition for multi-target tracking,'' in \emph{2015 IEEE Conference on
  Computer Vision and Pattern Recognition (CVPR)}, June 2015, pp. 5033--5041.

\bibitem{K4}
B.~W. Kernighan and S.~Lin, ``An efficient heuristic procedure for partitioning
  graphs,'' \emph{Bell system technical journal}, vol.~49, no.~2, pp. 291--307,
  1970.

\bibitem{ref34}
D.~Wilbers, C.~Merfels, and C.~Stachniss, ``Localization with sliding window
  factor graphs on third-party maps for automated driving.''

\bibitem{metric2}
M.~A. Brubaker, A.~Geiger, and R.~Urtasun, ``Lost! leveraging the crowd for
  probabilistic visual self-localization,'' in \emph{Proceedings of the IEEE
  Conference on Computer Vision and Pattern Recognition}, 2013, pp. 3057--3064.

\bibitem{ref38}
N.~Bodla, B.~Singh, R.~Chellappa, and L.~S. Davis, ``Soft-nms--improving object
  detection with one line of code,'' in \emph{Proceedings of the IEEE
  international conference on computer vision}, 2017, pp. 5561--5569.

\bibitem{ORB}
E.~Rublee, V.~Rabaud, K.~Konolige, and G.~R. Bradski, ``Orb: An efficient
  alternative to sift or surf.'' in \emph{ICCV}, vol.~11, no.~1.\hskip 1em plus
  0.5em minus 0.4em\relax Citeseer, 2011, p.~2.

\bibitem{sivic2003video}
J.~Sivic and A.~Zisserman, ``Video google: A text retrieval approach to object
  matching in videos,'' in \emph{null}.\hskip 1em plus 0.5em minus 0.4em\relax
  IEEE, 2003, p. 1470.

\bibitem{lowe2004distinctive}
D.~G. Lowe, ``Distinctive image features from scale-invariant keypoints,''
  \emph{International journal of computer vision}, vol.~60, no.~2, pp. 91--110,
  2004.

\bibitem{IBOW}
E.~Garcia-Fidalgo and A.~Ortiz, ``{iBoW-LCD: An Appearance-based Loop Closure
  Detection Approach using Incremental Bags of Binary Words},'' 2018.

\bibitem{galvez2012bags}
D.~G{\'a}lvez-L{\'o}pez and J.~D. Tardos, ``Bags of binary words for fast place
  recognition in image sequences,'' \emph{IEEE Transactions on Robotics},
  vol.~28, no.~5, pp. 1188--1197, 2012.

\bibitem{ref40}
\BIBentryALTinterwordspacing
J.~G. Mangelson, D.~Dominic, R.~M. Eustice, and R.~Vasudevan, ``Pairwise
  consistent measurement set maximization for robust multi-robot map merging,''
  in \emph{2018 {IEEE} International Conference on Robotics and Automation,
  {ICRA} 2018, Brisbane, Australia, May 21-25, 2018}, 2018, pp. 2916--2923.
  [Online]. Available: \url{https://doi.org/10.1109/ICRA.2018.8460217}
\BIBentrySTDinterwordspacing

\bibitem{shi1994good}
J.~Shi \emph{et~al.}, ``Good features to track,'' in \emph{1994 Proceedings of
  IEEE conference on computer vision and pattern recognition}.\hskip 1em plus
  0.5em minus 0.4em\relax IEEE, 1994, pp. 593--600.

\bibitem{ref10}
A.~Gawel, C.~Del~Don, R.~Siegwart, J.~Nieto, and C.~Cadena, ``X-view:
  Graph-based semantic multi-view localization,'' \emph{IEEE Robotics and
  Automation Letters}, vol.~3, no.~3, pp. 1687--1694, 2018.

\bibitem{seg1}
M.~Siam, C.~Jiang, S.~Lu, L.~Petrich, M.~Gamal, M.~Elhoseiny, and M.~Jagersand,
  ``Video object segmentation using teacher-student adaptation in a human robot
  interaction (hri) setting,'' in \emph{2019 International Conference on
  Robotics and Automation (ICRA)}.\hskip 1em plus 0.5em minus 0.4em\relax IEEE,
  2019, pp. 50--56.

\bibitem{ref27}
N.~Lee, S.~Ahn, and D.~Han, ``Amid: Accurate magnetic indoor localization using
  deep learning,'' \emph{Sensors}, vol.~18, no.~5, p. 1598, 2018.

\end{thebibliography}

\end{document}